\documentclass{article}

\PassOptionsToPackage{numbers, compress}{natbib}
\usepackage[final]{nips_2017}

\usepackage[utf8]{inputenc} 
\usepackage[T1]{fontenc}    
\usepackage{hyperref}       
\usepackage{url}            
\usepackage{booktabs}       
\usepackage{amsfonts}       
\usepackage{nicefrac}       
\usepackage{microtype}      
\usepackage{amsmath}
\usepackage{xspace}
\usepackage{subcaption}
\usepackage{graphicx}

\graphicspath{ {figures/} }

\newcommand{\sockeye}{\textsc{Sockeye}\xspace}
\newcommand{\mxnet}{\textsc{MXNet}\xspace}
\renewcommand{\vec}[1]{\boldsymbol{\mathbf{#1}}} 
\newcommand{\lvar}[1]{\mathit{#1}} 
\DeclareMathOperator*{\softmax}{softmax}
\DeclareMathOperator*{\argmax}{arg\,max}

\title{Image Captioning as Neural Machine \\ Translation Task in \sockeye}

\author{
  Loris Bazzani \\
  Amazon Research \\
  Core AI, Berlin \\
  \texttt{bazzanil@amazon.com}
  \And
  Tobias Domhan \\
  Amazon Research \\
  Machine Translation, Berlin \\
  \texttt{domhant@amazon.com}
  \And
  Felix Hieber \\
  Amazon Research \\
  Machine Translation, Berlin \\
  \texttt{fhieber@amazon.com}
}

\begin{document}

\maketitle

\begin{abstract}
Image captioning is an interdisciplinary research problem that stands between computer vision and natural language processing. 
The task is to generate a textual description of the content of an image.
The typical model used for image captioning is an encoder-decoder deep network, where the encoder captures the essence of an image while the decoder is responsible for generating a sentence describing the image.
Attention mechanisms can be used to automatically focus the decoder on parts of the image which are relevant to predict the next word.
In this paper, we explore different decoders and attentional models popular in neural machine translation, namely attentional recurrent neural networks, self-attentional transformers, and fully-convolutional networks, which represent the current state of the art of neural machine translation.
The image captioning module is available as part of \sockeye~\citep{hieber2017sockeye} at \url{https://github.com/awslabs/sockeye} which tutorial can be found at \url{https://awslabs.github.io/sockeye/image_captioning.html}.
\end{abstract}

\section{Introduction}\label{sec:introduction}
Image captioning is a relatively recent interdisciplinary research problem that stands between computer vision and natural language processing. 
The recent advances of both fields enables us to train models with good performance in practical applications.
We witnessed the huge wave of convolutional networks which revolutionized computer vision given their impressive results on image understanding using large annotated datasets~\cite{ILSVRC15}.
On the other hand, neural language models (\emph{e.g.}, sequence-to-sequence models for machine translation~\cite{sutskever2014sequence}) changed the way that natural language processing is done.

The task of image captioning is to generate a textual description of the content of an image.
Image captioning has multiple real-world applications: it can aid visually impaired users to obtain visual information and potentially interact by asking questions, it can be a way to interact with devices, it can be used to perform visual search using captions since text search is very optimized, among many others.

Image captioning requires a deep understanding of the content of the image that goes beyond object recognition and detection.
It is not only necessary to recognize most of the objects in the image, but also characterize their interaction, attach adjectives and compose a sentence that makes sense.
One possibility is to detect all the objects in an image, and use that to constrain the generation of the description~\cite{mitchell2012midge,karpathy2015deep}.
However, this requires to annotate the location of all the objects that appears in the vocabulary as well as associate them to the corresponding location in the sentence. 
This is clearly not scalable and requires a lot of annotation labor. 

More recently, solutions that use only image-text pairs are preferred to avoid overcome to the aforementioned problem.
In this scenario, the model has to learn the correspondences between part of the image and object label during training along with the grammar necessary to construct sentences. 
To achieve this goal, previous work~\cite{xu2015show,you2016image,lu2017knowing} proposed to learn models that are able to focus on the parts of the image that are responsible for a certain word in a description, i.e., attentional models.

A typical model used for image captioning shares many similarities with Neural Machine Translation (NMT) models: an encoder-decoder deep network, where the encoder captures the essence of an image while the decoder is responsible to generate the sentence.
For the encoder, deep convolutional networks trained for object recognition can be used as feature extractor. 
The decoder can be a Recurrent Neural Network (RNN), that takes as input the image features and the previous word, and predicts the next word. 
In practice, it has been observed that a single, fixed context for each word has some limitations~\cite{lu2017knowing}, because different parts of the image are related to different words. 
Therefore, attentional layers are adopted to relate different parts of the image to the respective words.

In this paper, we are exploring different decoder and attentional models for image captioning. 
Specifically, we decode the caption by using RNNs~\citep{Bahdanau:14,Luong:15}, transformers~\cite{transformer} and CNNs~\cite{convseq2seq} combined to different kinds of attentional mechanisms.

Moreover, we integrate image captioning into \sockeye~\citep{hieber2017sockeye}, an open-source sequence-to-sequence toolkit for NMT.
This is motivated by the fact that both NMT models and image captioning models follow the encoder-decoder structure.
Therefore, image captioning models can benefit from the features implemented in \sockeye: scalable training and inference for the most prominent encoder-decoder architectures in machine translation.
As outcome, we make our image-to-text module accessible to everyone.
Given the generality of the framework, it can be used to any image to sentence task, such as question generation in visual question answering tasks~\cite{Teney_2018_CVPR}.

\section{Related Works}\label{sec:related_works}

Following the characterization of the state of the art made in~\cite{lu2017knowing}, the literature is split in template-based approaches and neural-based approaches.
The first category includes the earlier attempts to solve image captioning, which consists of methods that fill slots in sentences based on detected objects, visual attributes and scenes, such as~\cite{farhadi2010every,kuznetsova2012collective,mitchell2012midge}.

More recently, neural-based approaches are taking over, because of the recent successes in image recognition and NMT.
These methods are inspired by the sequence-to-sequence encoder-decoder models that are dominant in machine translation nowadays~\cite{sutskever2014sequence}.
In NMT, the encoder takes as input a sequence in a source language and passes it to the decoder that outputs another sequence in a target language.
Image captioning can be seen as a image to text translation problem~\cite{karpathy2015deep}, where the source is not a sequence but an image.

Most works~\cite{vinyals2015show,chen2015mind,kiros2014multimodal} encode the image with the last fully-connected layer of a deep convolutional network, then the decoder can be either a feed-forward network~\cite{kiros2014multimodal} or an RNN~\cite{vinyals2015show,karpathy2015deep,lu2017knowing}.
However, a big problem of this strategy is that the spatial information is completely lost.
The works in~\cite{karpathy2015deep, anderson2018bottom} deal with this by detecting objects in the image. 

Attentional models~\cite{xu2015show,lu2017knowing,Luong:15} have been introduced to solve this problem.
The idea is to obtain an encoding of the image which is spatial variant, and train a model to decide which part of the image corresponds to each word of the target caption.
\cite{lu2017knowing} encodes images using the convolutional map of a ResNet, then projects it to a lower dimensionality. 
The attentional model is a feed-forward network that produces a value for each spatial component of the convolutional map (saliency map).
This information is used as context combined with the output of the decoder that is responsible of generating the next word.
\cite{lu2017knowing} proposes also a visual sentinel to deal with words that have no visual meaning (\emph{e.g.}, ``a'', ``the'', ...).
The work in~\cite{anderson2018bottom} extends the idea of attention on object detections instead of pixels in the convolutional map.
This work propose to extract features for each detected object which are then used by the attention module.

\section{Method}\label{sec:method}

The goal of image captioning is to model the probability distribution $p(Y|X; \vec{\theta})$, where $X$ is a source image, $Y=(y_1,...,y_m)$ is a target description, and  $\vec{\theta}$ is a parametrization of the chosen models.
Each $y_t$ is an integer id given by target vocabulary mapping, $\vec{V}_{\lvar{trg}}$, built from the training data tokens
 and represented as one-hot vectors $\vec{y}_t\in \{0,1\}^{|\vec{V}_{\lvar{trg}}|}$.
These are embedded into $e$-dimensional vector representations,  $\vec{E}_T\vec{y}_t$, using a learned embedding matrix $\vec{E}_T \in \mathbb{R}^{e\times|\vec{V}_{\lvar{trg}}|}$.

The probability can be factorized as follows:
\begin{align}
p(Y|I; \vec{\theta}) =  \prod_{t=1}^m p(y_t|Y_{1:t-1}, X; \vec{\theta}).
\end{align}
Learning reduces to finding the set of parameters that maximize the log likelihood:
\begin{align}
\vec{\theta}^* =  \argmax_{\vec{\theta}} \sum_{t=1}^m \log(p(y_t|Y_{1:t-1}, X; \vec{\theta})). \label{eq:optimization}
\end{align}

$p(y_t|Y_{1:t-1}, X; \vec{\theta})$ is parameterized via a softmax output layer over some decoder representation $\bar{\vec{h}}_t$:
 \begin{align}
 \label{eq:output}
 p(y_t|Y_{1:t-1}, X; \vec{\theta}) &= \softmax(\vec{W}_o\bar{\vec{h}}_t + \vec{b}_o), 
 \end{align}
where $\vec{W}_o$ scales to the dimension of the target vocabulary $\vec{V}_{\lvar{trg}}$.
Even though it is not explicit in the equation, the decoder representation $\bar{\vec{h}}_t$ depends on the image $X$.
Given the high dimensional nature of the image, it is necessary to encode it into a more meaningful and lower dimensional representation. 
In the next sections, we describe how the image is encoded and then how it is decoded to generate captions.
 
The maximization problem of Eq.~\ref{eq:optimization} is solved by optimizing the cross-entropy loss given Eq.~\ref{eq:output}.
This is straightforward to do since it is differentiable, however the cross-entropy loss might not correlate well with human judgment of machine-generated captions.
Other image captioning metrics such as CIDEr~\cite{vedantam2015cider} have been showed to better represent human perception.
Since they are not differentiable, reinforcement learning techniques can be used to solve the optimization problem (\emph{e.g.}, self-critical sequence training~\cite{rennie2017self}).
Although these technique have been showed to work well in practice, we postpone their integration with \sockeye to a future work.

\subsection{Image Encoder}\label{ssec:image_encoder}

The purpose of the image encoder is to project the image into a semantic feature space that is lower dimensionality than the original, full-resolution image.
For this task, ConvNets that are pre-trained for image recognition are used.
Since they are trained to recognize objects in images, it is very likely that they are activated in correspondence to object words in the context of captioning.

In particular, we follow the work in~\cite{lu2017knowing} that uses a ResNet-152~\cite{he2016deep} pre-trained on ImageNet.
We selected the last convolutional layer (namely, \texttt{stage4\_unit3\_conv3}), because it is the last layer of the network that retains the spatial information before the fully-connected layers.
This is important, since the decoder needs to be provided with the ability to correlate words to different parts of the image via attention, as we will see in the next section.

In our experiments, we also used features coming from object detections as in~\cite{anderson2018bottom}.
In that case we used ResNet-101 pre-trained on the Visual Genome dataset.
Each detected object is then represented by a mean-pooled convolutional feature from its region.

The resulting feature map is a matrix $\vec{F} = [ \vec{f}_1; \dots; \vec{f}_K ]$, where $K$ correspond to the spatial locations in the feature map or detections, and each vector $\vec{f}_k$ is a $d$-dimensional feature  ($2048$ here).
Moreover, it is useful to have a global image description $\vec{f}^g$ by average pooling such descriptors over the dimension $k$.
Since the feature dimensionality $d$ of the convolutional layer is often high, the ResNet features are projected to a lower dimension $d'<<d$ using a fully-connected layer ($512$ in this work):
\begin{align}
\vec{v}_k = ReLU(\vec{W}_{f}  \vec{f}_k), \quad \vec{v}^g = ReLU\left(\vec{W}_{g} \frac{1}{K} \sum_{k=1}^{K}{\vec{f}_k}\right),
\end{align}
composing the matrix $\vec{V} = [ \vec{v}_1; \dots; \vec{v}_K;  \vec{v}^g]$, which also contains the global image descriptor.

\subsection{Caption Decoders}\label{ssec:caption_decoders}

The decoder can be decomposed into 2 modules: 1) a temporal model that is able to encode the temporal information of the sequence of words, and 2) an attentional model that filters and selects the information coming from the encoder which is used as context to make predictions on the next word of the sequence.

In this work, we explore different possibilities for the temporal model as well as for the attentional model.
In particular, we exploit the similarity that between NMT models and the image captioning ones, by testing different state-of-the-art decoders for NMT, namely: attentional recurrent neural networks, self-attentional transformers, and fully-convolutional networks.
We will briefly describe those models in the following.
Please refer to~\cite{hieber2017sockeye} for their description in the NMT domain.

\paragraph{Attentional RNN~\citep{Bahdanau:14,Luong:15}.}
The decoder is defined as:
\begin{align}
\vec{h}_t = f_{dec} ([\vec{E}_T \vec{y}_{t-1}; \vec{\bar{h}}_{t-1}], \vec{h}_{t-1}), \label{eq:RNN}
\end{align}
where $f_{dec}$ is a (multi-layer) RNN, $\vec{h}_{t-1}$ is the previous hidden representation of the RNN, and $\vec{\bar{h}}_{t-1}$ is the image-dependent attentional vector.

Attention is computed by considering the input image representation as well as the current word hidden representation as follows:
\begin{align}
score(\vec{v}_k, \vec{h}_t) &= \vec{w}_h^T tanh (\vec{W}_v \vec{v}_k + \vec{W}_h \vec{h}_t), \\
  \alpha_{kt} &= \softmax(score(\vec{v}_k, \vec{h}_t) ).
\end{align}
The context vector is built by the sum of hidden vectors weighted by the attentional score: $\vec{c}_t = \sum_{k=1}^{K}{\alpha_{kt} \vec{h}_t}$.
This kind of attention is often called Multi-Layer Perceptron (MLP) attention, given that the score is computed with MLP-style model. 
Another common type of attentional model can be the dot product between $\vec{v}_k$ and $ \vec{h}_t$, or the multi-head attention used by the transformer (see below).

Attention and decoder are combined to build the image-dependent attentional vector as:
\begin{align}
\vec{\bar{h}}_{t} = tanh( \vec{W}_{\bar{h}} [\vec{h}_{t}; \vec{c}_t ]),
\end{align}
which is then used to predict the next word as in Eq.~\ref{eq:output}.

In Eq.~\ref{eq:RNN}, we can optionally concatenate the global image descriptor to the input of the RNN: $[\vec{E}_T \vec{y}_{t-1}; \vec{\bar{h}}_{t-1}; \vec{v}^g]$.
This is similar to~\cite{lu2017knowing}, with the only difference that their model does not use $\vec{\bar{h}}_{t-1}$ as input of the RNN.

\paragraph{Self-attentional Transformer~\cite{transformer}.}
The transformer model uses attention to replace recurrent dependencies, making the representation at a certain time step independent from the other time steps. 
This allows for parallelization of the computation for all time steps in the decoder at training time.

Since the recurrent dependencies are removed, time needs to be explicitly encoded as positional information in a sequence as: $\vec{E}_T \vec{x}_i + \vec{e}_{pos,i}$, where $ \vec{e}_{pos,i}$ is the positional embedding at position $i$, which can be learned or fixed~\cite{transformer}.

The transformer has a self-attention mechanism which is a generalization of \citep{Luong:15} and is defined as: 
 \begin{align}
     \vec{C}_u = \softmax \left( \frac{\vec{Q}\vec{W}^Q_u (\vec{K}\vec{W}^K_u)^\top}{\sqrt{d_u}} \right) \vec{L}\vec{W}^L_u, \label{eq:selfattention}
 \end{align}
where $\vec{C}_u$ is the context matrix produced by a head, $\vec{Q} \in \mathbb{R}^{n \times d}$ is a query matrix, $\vec{K} \in \mathbb{R}^{n \times d}$ is a key matrix, $\vec{L} \in \mathbb{R}^{n \times d}$ is a value matrix, the $\vec{W}_u$s are projections, $n$ is the number of hidden states and $d$ denotes the number of hidden units.
The final context matrix is given by concatenating the heads, followed by a linear transformation: $\vec{C} = [\vec{C}_1; .\dots; \vec{C}_h] \vec{W}_O$.

The second subnetwork is a feed-forward network with ReLU activation defined as
\begin{align}
FFN(\vec{x}) = ReLU(\vec{x} \vec{W}_1 + \vec{b}_1) \vec{W}_2 + \vec{b}_2.
\end{align}
Each sublayer, self-attention and feed-forward network, is followed by a post-processing stack of dropout and layer normalization~\cite{layernorm}.

The sequence of operations of a single decoder block is:
\begin{align*}
\small
\text{Self-attention} \rightarrow \text{Post-process} \rightarrow \text{Encoder attention} \rightarrow \text{Post-process} \rightarrow \text{Feed-forward} \rightarrow \text{Post-process}
\end{align*}
Multiple blocks are stacked to form the full decoder network and the representation of the last block is fed into the output layer.
For the decoder, self-attention is restricted to $\vec{Q} = \vec{K} = \vec{L} = \vec{H}$ in Eq.~\ref{eq:selfattention}, where $\vec{H}$ is the matrix of hidden states $\vec{h}_t$ (input embeddings in the first layer).
For the encoder, we have  $\vec{Q} = \vec{H}$ and $\vec{K} = \vec{L} = \vec{V}$, where $\vec{V}$ contains the image encoder states as described in Sec,~\ref{ssec:image_encoder}.

\paragraph{Fully-convolutional Network~\cite{convseq2seq}.}
As for the transformer network there are no recurrent depedencies, therefore the positional embedding is used.
Then, a sequence of stacked convolutional layers~\cite{Aneja_2018_CVPR} is applied to the embeddings, \emph{i.e.}, the inputs of the decoder.

The output of each convolutional layer is fed to the attentional mechanism described for the transformer with a single attention head.
In particular, $\vec{Q} = \vec{H_c}$ and $\vec{K} = \vec{L} = \vec{V}$, where $\vec{H_c}$ is the output of the convolution and $\vec{V}$ is the image encoder state.
The decoder hidden state is a residual combination with the context vector, which is then fed to the output layer.

\section{Experiments}\label{sec:experiments}

The experiments were carried out on two popular datasets for image captioning: Flickr30k~\cite{plummer2015flickr30k,young2014image} and MS-COCO~\cite{lin2014microsoft}.
Flickr30k contains $31,783$ images, each of which has $5$ captions on average.
We use the splits provided in~\cite{johnson2016densecap} which contains $1,000$ images for validation and $1,000$ images for testing.

MS-COCO is a bigger dataset containing $82,783$, $40,504$ and $40,775$ images for training, validation and test, respectively.
$5$ captions on average are provided for each image.
We use the splits provided in~\cite{johnson2016densecap} which contains $5,000$ images for validation and $5,000$ images for testing, taken from the original validation set.
The remainder of the validation set ($30,504$ images) is merged to the training set, producing a set of $113,287$ images.

The results are reported in terms of standard metrics for machine translation and image captioning\footnote{We use the evaluation code at \url{https://github.com/tylin/coco-caption}.}, such as BLEU at different N-grams (B@N)~\cite{papineni2002bleu}, METEOR~\cite{denkowski2014meteor} and CIDEr~\cite{vedantam2015cider}.

\paragraph{Implementation Details.}
The models and experiments are implemented in \mxnet in the \sockeye framework.
The ResNet is used as feature extractor, therefore not finetuned, in our experiments. 
Training is performed using the Adam optimizer with batch size of $64$, initial learning rates of $0.0005$ and $0.0003$ and absolute gradient clipping of $1.0$.
The learning rate is reduced by a factor $0.9$ when the validation perplexity does not change for $3$ iterations.
The RNN is an LSTM with $512$ units, the same number as for the transformers. 
During inference, we perform beam search with size of $3$ and $5$.
For Flickr30k only, we do not consider words in the vocabulary with frequency less than $5$.

\subsection{Variants of the Proposed Model}\label{ssec:variants}

\begin{table}
  \caption{Which decoder is the best? Dataset: MS-COCO.}
  \label{tab:decoder_fight}
  \footnotesize
  \centering
  \renewcommand{\arraystretch}{1.2}
  \begin{tabular}{r @{\hskip6pt} c @{\hskip6pt} c @{\hskip6pt} c @{\hskip6pt} c @{\hskip6pt} c @{\hskip6pt} c @{\hskip6pt}}
  \toprule
	Method & B@1 & B@2 & B@3 & B@4 & METEOR & CIDEr \\  \hline
ARNN & 0.672	& 0.496	& 0.364	& 0.271	& \textbf{0.246}	& \textbf{0.882} \\
SAT &  \textbf{0.677} & \textbf{0.502}	& \textbf{0.371}	& \textbf{0.278}	& 0.243 & \textbf{0.882} \\
FCN & 0.661 & 0.482 & 0.347	& 0.254 & 0.231 & 0.803 \\
  \bottomrule
  \end{tabular}
\end{table}

In this section, we evaluate the variants of the proposed model using the  MS-COCO dataset.
First of all, we want to establish which decoder architecture is the best for image captioning.
We did a comparative study among the three architectures presented in Sec.~\ref{ssec:caption_decoders}, namely Attentional Recurrent Neural Networks (ARNN), Self-Attentional Transformers (SAT), and Fully-Convolutional Networks (FCN).
The results are reported in Table~\ref{tab:decoder_fight}.
It is clear that FCN gives the worst results out of the three models.
On the other hand, SAT compares favorably with ARNN with a small gap between them, thus there is not clear winner.

Table~\ref{tab:attention_fight} reports the results for the different kind of attentional models that can be used in combination with ARNN as described in Sec.~\ref{ssec:caption_decoders}.
In particular, we tested the following variants:
\begin{enumerate}
\item \emph{Without}: attention is removed, instead we use the global image descriptor as context vector.
\item \emph{Dot}: dot product between the image representation and the hidden representation of the RNN.
\item \emph{Multihead-8}: multihead attention with 8 heads~\cite{transformer}.
\item \emph{MLP}: multi-layer perceptron attention.
\end{enumerate}
One can notice that dot attention gives poor results, and this might due to the fact that the image representations are not properly normalized before being fed to the output layer.
In fact, Multihead-8 and MLP are the best attentional models.
As expected, attention is required for image captioning, although the gap with respect to the MLP attention is narrow.

\begin{table}
  \caption{Which attentional model gives the best results? Model: ARNN. Dataset: MS-COCO.}
  \label{tab:attention_fight}
  \footnotesize
  \centering
  \renewcommand{\arraystretch}{1.2}
  \begin{tabular}{r @{\hskip6pt} c @{\hskip6pt} c @{\hskip6pt} c @{\hskip6pt} c @{\hskip6pt} c @{\hskip6pt} c @{\hskip6pt}}
  \toprule
	Method & B@1 & B@2 & B@3 & B@4 & METEOR & CIDEr \\  \hline
Without & 0.666 & 	0.489 & 0.357 & 	0.265 & 0.242	& 0.861 \\
Dot &  0.497 & 0.303	& 0.188	& 0.122	& 0.148	& 0.306 \\
Multihead-8 & 0.671 & \textbf{0.496}	& \textbf{0.365}	& \textbf{0.273}	& \textbf{0.244}	& \textbf{0.882}\\
MLP &  \textbf{0.672}	& \textbf{0.496}	& 0.364	& 0.271	& 0.246	& \textbf{0.882} \\
  \bottomrule
  \end{tabular}
\end{table}

Now that we discovered that 1) both ARNN and SAT give good results and 2) we need attention, we can explore other extensions of these models. 
Table~\ref{tab:optimize_arnn_sat} shows that both ARNN and SAT improve by:
\begin{enumerate}
\item Reducing the beam size from 5 to 3 (second row),
\item Concatenating the image global descriptor to the word embedding  (third row),
\item Adding weight normalization (last row).
\end{enumerate}

\begin{table}
  \caption{Incremental addition of features for ARNN and SAT. Dataset: MS-COCO.}
  \label{tab:optimize_arnn_sat}
  \scriptsize
  \centering
  \renewcommand{\arraystretch}{1.2}
  \begin{tabular}{l @{\hskip6pt} c @{\hskip6pt} c @{\hskip6pt} c @{\hskip6pt} c @{\hskip6pt} c @{\hskip6pt} c  c @{\hskip6pt} c @{\hskip6pt} c @{\hskip6pt} c @{\hskip6pt} c @{\hskip6pt} c @{\hskip6pt}}
  \toprule
  	& \multicolumn{6}{c}{ARNN} & \multicolumn{6}{c}{SAT}\\ \cmidrule(r){2-7} \cmidrule(l){8-13}
	Method & B@1 & B@2 & B@3 & B@4 & METEOR & CIDEr & B@1 & B@2 & B@3 & B@4 & METEOR & CIDEr \\  \hline
Original & 0.672	& 0.496	& 0.364	& 0.271	& 0.246	& 0.882 & 0.677 & 0.502	& 0.371	& 0.278	& 0.243 & 0.882	 \\
+ beam size = 3 &  0.683 &	0.510 &	0.378 &	0.283 &	0.248 &	0.919 & 0.687	& 0.513	& 0.379	& 0.283	& 0.248	& 0.922	 \\
+ concat image global &  0.688 & 	0.514 &	0.380 &	0.283 &	0.250 &	0.924  &	- & - & - & - & - & - \\
+ weight norm & 0.692 &	0.519 &	0.385 &	0.287 &	0.250 &	0.934  &	0.687 	& 0.513	& 0.380	& 0.282	& 0.248	& 0.923\\
  \bottomrule
  \end{tabular}
\end{table}

We run many other experiments that were not significantly changing the results.
We found that in ARNN replacing LSTM units with GRU units or increasing the number of hidden units from $512$ to $1024$ give comparable results (small difference of 0.002 B@4).
Using a deep, 8-layer ARNN with residual connections is decreasing the B@4 of 0.007 points.
A similar conclusion can be drawn by comparing a deep, 2-layer transformer with its shallow counterpart (difference of 0.006 B@4).
Finally, we tried to finetune the image encoder, however this degraded the performance.
As discussed in~\cite{lu2017knowing}, finetuning should be done carefully by activating it only in the last phase of training when the validation error stabilizes.
It is worth noting that we did not investigate and dive deep on this option, so we leave it as future work.

\subsection{Comparison with the State of the Art}\label{ssec:state_of_the_art}

In this section, we compare our best model with the current state of the art.
As showed by our analysis in the previous section, our best model is an ARNN with $512$ hidden units, MLP attention, concatenation of the image global descriptor, weight normalization and beam size = 3 for inference.
We also trained our best model on top of object detection features as in~\citep{anderson2018bottom}.

\begin{table}
  \caption{Comparison with the state of the art. Datasets: Flickr30k and MS-COCO.}
  \label{tab:soa}
  \scriptsize
  \centering
  \renewcommand{\arraystretch}{1.2}
  \begin{tabular}{l @{\hskip6pt} c @{\hskip6pt} c @{\hskip6pt} c @{\hskip6pt} c @{\hskip6pt} c @{\hskip6pt} c  c @{\hskip6pt} c @{\hskip6pt} c @{\hskip6pt} c @{\hskip6pt} c @{\hskip6pt} c @{\hskip6pt}}
  \toprule
  	& \multicolumn{6}{c}{Flickr30k} & \multicolumn{6}{c}{MS-COCO}\\ \cmidrule(r){2-7} \cmidrule(l){8-13}
	Method & B@1 & B@2 & B@3 & B@4 & METEOR & CIDEr & B@1 & B@2 & B@3 & B@4 & METEOR & CIDEr \\  \hline
    \begin{tabular}{r} DeepVS \citep{karpathy2015deep} \end{tabular} & 0.573 & 0.369 & 0.240 & 0.157 & 0.153 & 0.247 & 0.625 & 0.450 & 0.321 & 0.230 & 0.195 & 0.660 \\
    \begin{tabular}{r} LRCN \citep{donahue2015long} \end{tabular} & 0.587 & 0.391 & 0.251 & 0.165 & - & - & 0.669 & 0.489 & 0.350 & 0.249 & - & - \\
    \begin{tabular}{r} Hard-attention \citep{xu2015show} \end{tabular} & 0.669 & 0.439 & 0.296 & 0.199 & 0.185 & - & 0.718 & 0.504 & 0.357 & 0.250 & 0.230 & - \\
    \begin{tabular}{r} ATT-FCN \citep{you2016image} \end{tabular} & 0.647 & 0.460 & 0.324 & 0.230 & 0.189 & - & 0.709 & 0.537 & 0.402 & 0.304 & 0.243 & - \\
    \begin{tabular}{r} MSM \citep{yao2017boosting} \end{tabular} & - & - & - & - & - & - & 0.730 & 0.565 & 0.429 & 0.325 & 0.251 & 0.986 \\  
    \begin{tabular}{r} Spatial \citep{lu2017knowing} \end{tabular} & 0.644 & 0.462 & 0.327 & 0.231 & 0.202 & 0.493 &  0.734 & 0.566 & 0.418 & 0.304 & 0.257 & 1.029 \\
    \begin{tabular}{r} Adaptive \citep{lu2017knowing} \end{tabular} & 0.677 & 0.494 & 0.354 & 0.251 & 0.204 & 0.531 & 0.742 & \textbf{0.580} & 0.439 & 0.332 & 0.266 & 1.085 \\ 
    \begin{tabular}{r} Up-down \citep{anderson2018bottom} \end{tabular} & - & - & - & - & - & - & 0.766 & - & - & 0.340 & 0.265 & 1.111 \\    
    \begin{tabular}{r} Up-down CIDEr \citep{anderson2018bottom} \end{tabular} & - & - & - & - & - & - & \textbf{0.798} & - & - & \textbf{0.363} & \textbf{0.277} & \textbf{1.201} \\    
     \hline
    \begin{tabular}{r} Our Best \end{tabular}  & 0.625 &	0.437 &	0.300 &	0.205 &	0.193 &	0.453 & 0.692 &	0.519 &	0.385 &	0.287 &	0.250 &	0.934 \\
    \begin{tabular}{r} Our Best w/ Detector \end{tabular}  & \textbf{0.681} &	\textbf{0.507} &	\textbf{0.374} &	\textbf{0.274} &	\textbf{0.218} &	\textbf{0.594} & 0.739 &	0.577 &	\textbf{0.443}  &	0.339 &	0.276 &	1.098 \\
  \bottomrule
  \end{tabular}
\end{table}

Table~\ref{tab:soa} reports the current state of the art for image captioning along with our best method from the previous section (Our Best) and when using object detection features as in~\citep{anderson2018bottom} (Our best w/ Detector in Table~\ref{tab:soa}).
It is clear that when using object detection features, we obtain a significant improvement of about $>3\%$ on all metrics and datasets.
This model is better of all the paper excluding~\citep{anderson2018bottom} (Up-down in Table~\ref{tab:soa}) which still is the best especially because it optimizes CIDEr with reinforcement learning.

\begin{figure}[h!]
    \centering
    \includegraphics[width=\textwidth]{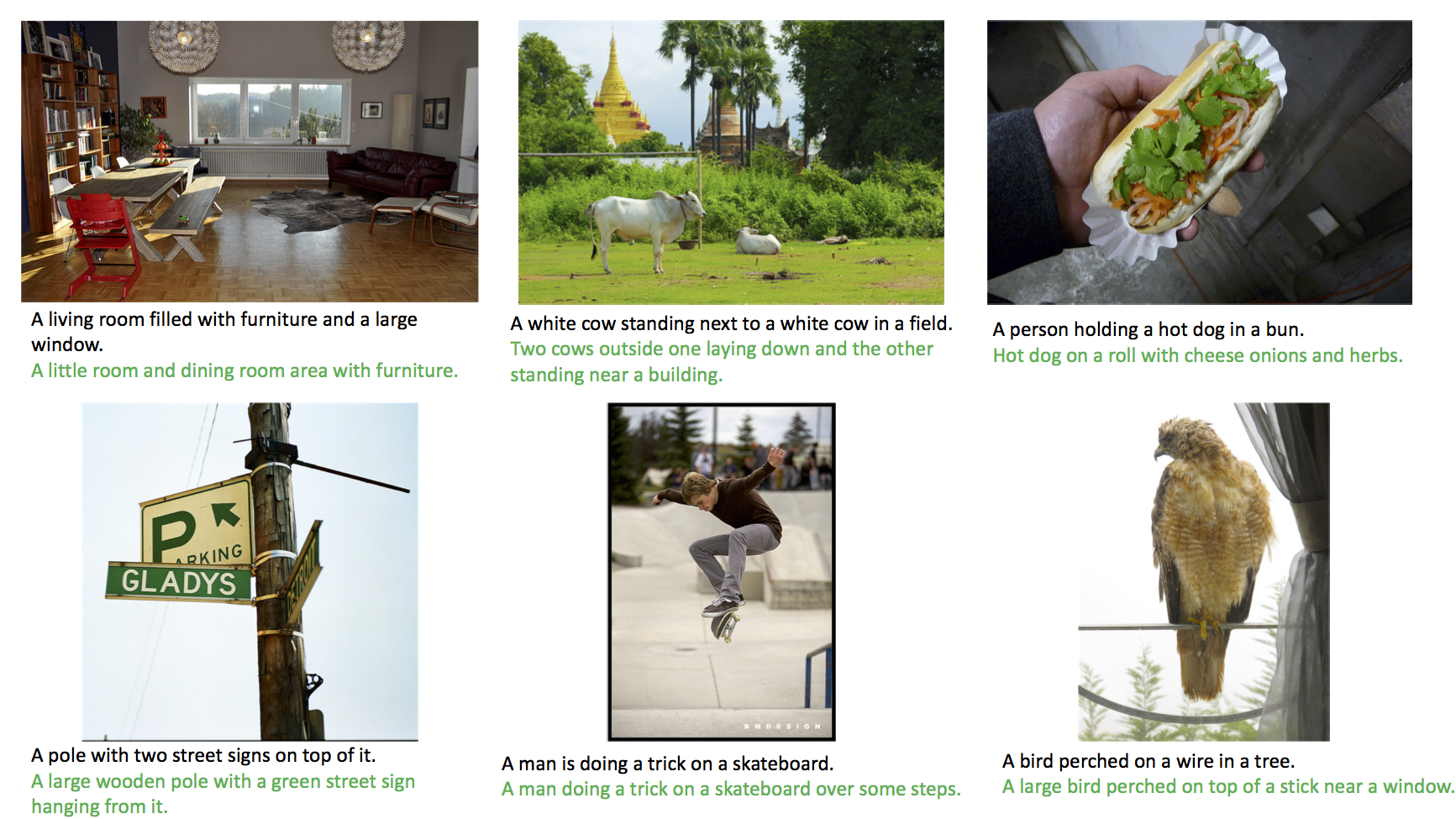}
    \includegraphics[width=\textwidth]{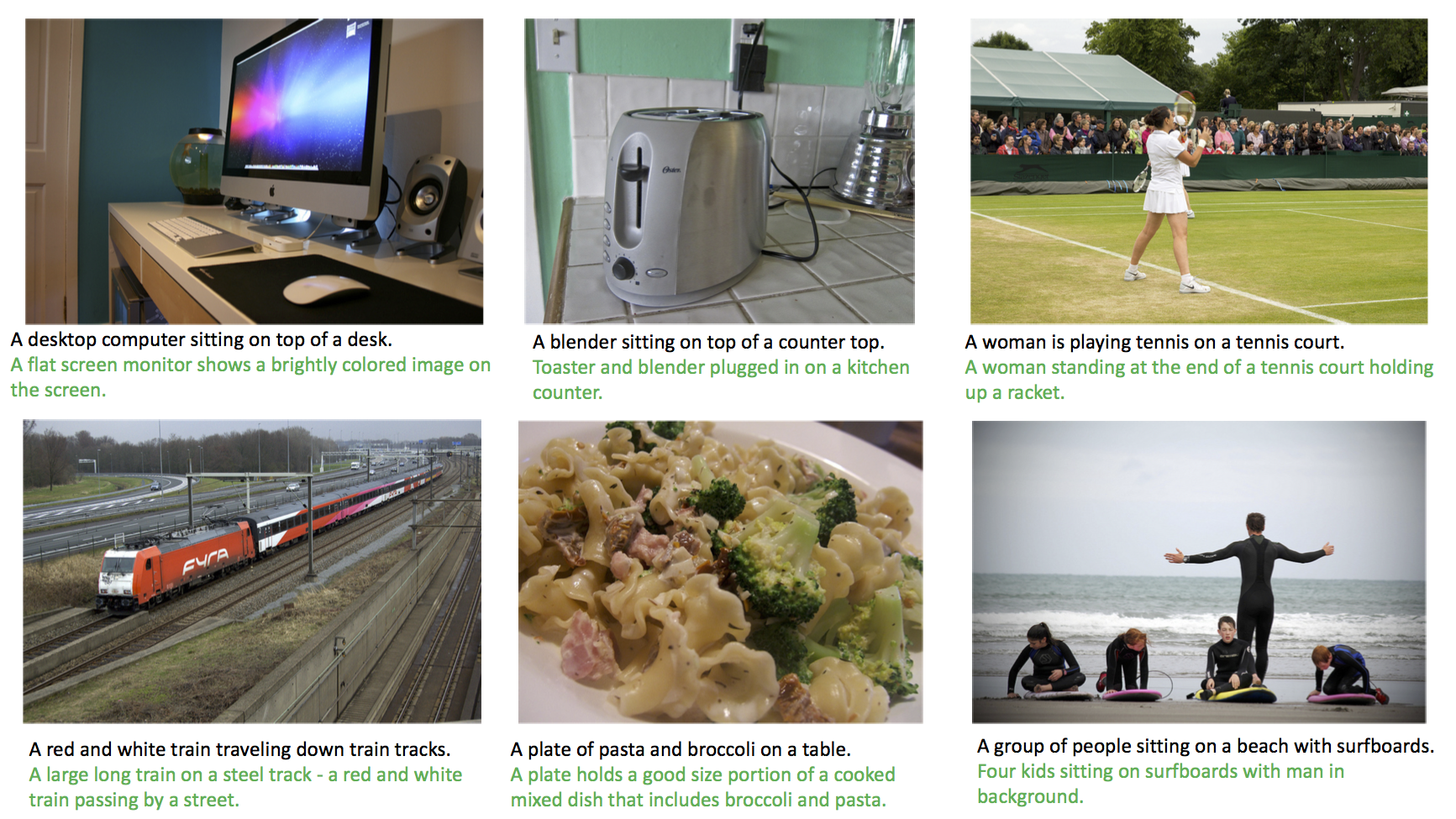}
    \caption{Visualizations for MS-COCO. Predictions and ground truth sentences are in black and green text, respectively. Better visualized zoomed on a computer monitor. \\
    { \small List of images and attribution: 
\href{https://www.flickr.com/photos/clg20171/6326677826/in/photostream/}{Living room} by \href{https://www.flickr.com/photos/clg20171/}{clg20171}; 
\href{https://www.flickr.com/photos/104284854@N07/10096943865/}{Cows and Inwa Temples}  by \href{https://www.flickr.com/photos/104284854@N07/}{KX Studio}; 
\href{https://www.flickr.com/photos/rexroof/4430471591/}{Asiadog's VINH}  by \href{https://www.flickr.com/photos/rexroof/}{Rex Roof}; 
\href{https://www.flickr.com/photos/stevensnodgrass/5744540082/}{For You, Gladys}  by \href{https://www.flickr.com/photos/stevensnodgrass/}{Steve Snodgrass}; 
\href{https://www.flickr.com/photos/30904483@N06/2890964329/}{Double Kickflip}  by \href{https://www.flickr.com/photos/30904483@N06/}{Brandon McKay}; 
\href{https://www.flickr.com/photos/gazeronly/6009338149/}{red tail}  by \href{https://www.flickr.com/photos/gazeronly/}{torbakhopper}; 
\href{https://www.flickr.com/photos/declanjewell/5812924771/}{New iMac}  by \href{https://www.flickr.com/photos/declanjewell/}{DeclanTM}; 
\href{https://www.flickr.com/photos/ex_magician/5926562644/}{Toaster}  by \href{https://www.flickr.com/photos/ex_magician/}{Michael (a.k.a. moik) McCullough}; 
\href{https://www.flickr.com/photos/ebby-rebby/9257068408/}{Wimbledon 2013 invitational doubles}  by \href{https://www.flickr.com/photos/ebby-rebby/}{Emma}; 
\href{https://www.flickr.com/photos/89246112@N00/8645262251/}{De FYRA-Variatie.}  by \href{https://www.flickr.com/photos/89246112@N00/}{Bart}; 
\href{https://www.flickr.com/photos/wordridden/1537200374/}{Creamy pasta}  by \href{https://www.flickr.com/photos/wordridden/}{WordRidden}; 
\href{https://www.flickr.com/photos/fotologic/6038911779/}{Learning to surf}  by \href{https://www.flickr.com/photos/fotologic/}{fotologic}. 
All images  are licensed under \href{http://creativecommons.org/licenses/by/2.0/}{CC BY 2.0}.} }
    \label{fig:viz}
\end{figure}

We show some qualitative results of the two datasets in Figure~\ref{fig:viz}.
The figures show the original image along with the predicted caption in white background and the set of ground truth descriptions provided by the annotators in green background.

\section{Conclusions}

In this work, we explored different decoders and attentional models for image captioning, and showed that the attentional RNN is slightly better than the self-attention transformer.
We were able to obtaining competitive results when comparing with the state of the art.
The next step is to close the gap between our method and~\citep{anderson2018bottom} by implementing reinforcement learning to optimize non-differentiable image captioning metrics.

\paragraph{Acknowledgments.} We thank Matt Post, Hagen Fuerstenau, Michael Donoser and David Vilar Torres for helping in brainstorming ideas, designing code and experiments, giving feedback on the code and on this technical report.

\clearpage
\bibliographystyle{plainnat}
\bibliography{bibliography}

\begin{thebibliography}{32}
\providecommand{\natexlab}[1]{#1}
\providecommand{\url}[1]{\texttt{#1}}
\expandafter\ifx\csname urlstyle\endcsname\relax
  \providecommand{\doi}[1]{doi: #1}\else
  \providecommand{\doi}{doi: \begingroup \urlstyle{rm}\Url}\fi

\bibitem[Anderson et~al.(2018)Anderson, He, Buehler, Teney, Johnson, Gould, and
  Zhang]{anderson2018bottom}
Peter Anderson, Xiaodong He, Chris Buehler, Damien Teney, Mark Johnson, Stephen
  Gould, and Lei Zhang.
\newblock Bottom-up and top-down attention for image captioning and visual
  question answering.
\newblock In \emph{CVPR}, volume~3, page~6, 2018.

\bibitem[Aneja et~al.(2018)Aneja, Deshpande, and Schwing]{Aneja_2018_CVPR}
Jyoti Aneja, Aditya Deshpande, and Alexander~G. Schwing.
\newblock Convolutional image captioning.
\newblock In \emph{The IEEE Conference on Computer Vision and Pattern
  Recognition (CVPR)}, June 2018.

\bibitem[Ba et~al.(2016)Ba, Kiros, and Hinton]{layernorm}
Jimmy~Lei Ba, Jamie~Ryan Kiros, and Geoffrey~E. Hinton.
\newblock Layer normalization.
\newblock \emph{CoRR}, abs/1607.06450, 2016.

\bibitem[Bahdanau et~al.(2014)Bahdanau, Cho, and Bengio]{Bahdanau:14}
Dzmitry Bahdanau, Kyunghyun Cho, and Yoshua Bengio.
\newblock Neural machine translation by jointly learning to align and
  translate.
\newblock \emph{CoRR}, abs/1409.0473, 2014.

\bibitem[Chen and Lawrence~Zitnick(2015)]{chen2015mind}
Xinlei Chen and C~Lawrence~Zitnick.
\newblock Mind's eye: A recurrent visual representation for image caption
  generation.
\newblock In \emph{Proceedings of the IEEE conference on computer vision and
  pattern recognition}, pages 2422--2431, 2015.

\bibitem[Denkowski and Lavie(2014)]{denkowski2014meteor}
Michael Denkowski and Alon Lavie.
\newblock Meteor universal: Language specific translation evaluation for any
  target language.
\newblock In \emph{Proceedings of the ninth workshop on statistical machine
  translation}, pages 376--380, 2014.

\bibitem[Donahue et~al.(2015)Donahue, Anne~Hendricks, Guadarrama, Rohrbach,
  Venugopalan, Saenko, and Darrell]{donahue2015long}
Jeffrey Donahue, Lisa Anne~Hendricks, Sergio Guadarrama, Marcus Rohrbach,
  Subhashini Venugopalan, Kate Saenko, and Trevor Darrell.
\newblock Long-term recurrent convolutional networks for visual recognition and
  description.
\newblock In \emph{Proceedings of the IEEE conference on computer vision and
  pattern recognition}, pages 2625--2634, 2015.

\bibitem[Farhadi et~al.(2010)Farhadi, Hejrati, Sadeghi, Young, Rashtchian,
  Hockenmaier, and Forsyth]{farhadi2010every}
Ali Farhadi, Mohsen Hejrati, Mohammad~Amin Sadeghi, Peter Young, Cyrus
  Rashtchian, Julia Hockenmaier, and David Forsyth.
\newblock Every picture tells a story: Generating sentences from images.
\newblock In \emph{European conference on computer vision}, pages 15--29.
  Springer, 2010.

\bibitem[Gehring et~al.(2017)Gehring, Auli, Grangier, Yarats, and
  Dauphin]{convseq2seq}
Jonas Gehring, Michael Auli, David Grangier, Denis Yarats, and Yann~N. Dauphin.
\newblock Convolutional sequence to sequence learning.
\newblock \emph{CoRR}, abs/1705.03122, 2017.

\bibitem[He et~al.(2016)He, Zhang, Ren, and Sun]{he2016deep}
Kaiming He, Xiangyu Zhang, Shaoqing Ren, and Jian Sun.
\newblock Deep residual learning for image recognition.
\newblock In \emph{Proceedings of the IEEE conference on computer vision and
  pattern recognition}, pages 770--778, 2016.

\bibitem[Hieber et~al.(2017)Hieber, Domhan, Denkowski, Vilar, Sokolov, Clifton,
  and Post]{hieber2017sockeye}
Felix Hieber, Tobias Domhan, Michael Denkowski, David Vilar, Artem Sokolov, Ann
  Clifton, and Matt Post.
\newblock Sockeye: A toolkit for neural machine translation.
\newblock \emph{arXiv preprint arXiv:1712.05690}, 2017.

\bibitem[Johnson et~al.(2016)Johnson, Karpathy, and
  Fei-Fei]{johnson2016densecap}
Justin Johnson, Andrej Karpathy, and Li~Fei-Fei.
\newblock Densecap: Fully convolutional localization networks for dense
  captioning.
\newblock In \emph{Proceedings of the IEEE Conference on Computer Vision and
  Pattern Recognition}, pages 4565--4574, 2016.

\bibitem[Karpathy and Fei-Fei(2015)]{karpathy2015deep}
Andrej Karpathy and Li~Fei-Fei.
\newblock Deep visual-semantic alignments for generating image descriptions.
\newblock In \emph{Proceedings of the IEEE conference on computer vision and
  pattern recognition}, pages 3128--3137, 2015.

\bibitem[Kiros et~al.(2014)Kiros, Salakhutdinov, and
  Zemel]{kiros2014multimodal}
Ryan Kiros, Ruslan Salakhutdinov, and Rich Zemel.
\newblock Multimodal neural language models.
\newblock In \emph{International Conference on Machine Learning}, pages
  595--603, 2014.

\bibitem[Kuznetsova et~al.(2012)Kuznetsova, Ordonez, Berg, Berg, and
  Choi]{kuznetsova2012collective}
Polina Kuznetsova, Vicente Ordonez, Alexander~C Berg, Tamara~L Berg, and Yejin
  Choi.
\newblock Collective generation of natural image descriptions.
\newblock In \emph{Proceedings of the 50th Annual Meeting of the Association
  for Computational Linguistics: Long Papers-Volume 1}, pages 359--368.
  Association for Computational Linguistics, 2012.

\bibitem[Lin et~al.(2014)Lin, Maire, Belongie, Hays, Perona, Ramanan,
  Doll{\'a}r, and Zitnick]{lin2014microsoft}
Tsung-Yi Lin, Michael Maire, Serge Belongie, James Hays, Pietro Perona, Deva
  Ramanan, Piotr Doll{\'a}r, and C~Lawrence Zitnick.
\newblock Microsoft coco: Common objects in context.
\newblock In \emph{European conference on computer vision}, pages 740--755.
  Springer, 2014.

\bibitem[Lu et~al.(2017)Lu, Xiong, Parikh, and Socher]{lu2017knowing}
Jiasen Lu, Caiming Xiong, Devi Parikh, and Richard Socher.
\newblock Knowing when to look: Adaptive attention via a visual sentinel for
  image captioning.
\newblock In \emph{Proceedings of the IEEE Conference on Computer Vision and
  Pattern Recognition (CVPR)}, volume~6, 2017.

\bibitem[Luong et~al.(2015)Luong, Pham, and Manning]{Luong:15}
Thang Luong, Hieu Pham, and Christopher~D. Manning.
\newblock Effective approaches to attention-based neural machine translation.
\newblock In \emph{EMNLP}, 2015.

\bibitem[Mitchell et~al.(2012)Mitchell, Han, Dodge, Mensch, Goyal, Berg,
  Yamaguchi, Berg, Stratos, and Daum{\'e}~III]{mitchell2012midge}
Margaret Mitchell, Xufeng Han, Jesse Dodge, Alyssa Mensch, Amit Goyal, Alex
  Berg, Kota Yamaguchi, Tamara Berg, Karl Stratos, and Hal Daum{\'e}~III.
\newblock Midge: Generating image descriptions from computer vision detections.
\newblock In \emph{Proceedings of the 13th Conference of the European Chapter
  of the Association for Computational Linguistics}, pages 747--756.
  Association for Computational Linguistics, 2012.

\bibitem[Papineni et~al.(2002)Papineni, Roukos, Ward, and
  Zhu]{papineni2002bleu}
Kishore Papineni, Salim Roukos, Todd Ward, and Wei-Jing Zhu.
\newblock Bleu: a method for automatic evaluation of machine translation.
\newblock In \emph{Proceedings of the 40th annual meeting on association for
  computational linguistics}, pages 311--318. Association for Computational
  Linguistics, 2002.

\bibitem[Plummer et~al.(2015)Plummer, Wang, Cervantes, Caicedo, Hockenmaier,
  and Lazebnik]{plummer2015flickr30k}
Bryan~A Plummer, Liwei Wang, Chris~M Cervantes, Juan~C Caicedo, Julia
  Hockenmaier, and Svetlana Lazebnik.
\newblock Flickr30k entities: Collecting region-to-phrase correspondences for
  richer image-to-sentence models.
\newblock In \emph{Computer Vision (ICCV), 2015 IEEE International Conference
  on}, pages 2641--2649. IEEE, 2015.

\bibitem[Rennie et~al.(2017)Rennie, Marcheret, Mroueh, Ross, and
  Goel]{rennie2017self}
Steven~J Rennie, Etienne Marcheret, Youssef Mroueh, Jarret Ross, and Vaibhava
  Goel.
\newblock Self-critical sequence training for image captioning.
\newblock In \emph{CVPR}, volume~1, page~3, 2017.

\bibitem[Russakovsky et~al.(2015)Russakovsky, Deng, Su, Krause, Satheesh, Ma,
  Huang, Karpathy, Khosla, Bernstein, Berg, and Fei-Fei]{ILSVRC15}
Olga Russakovsky, Jia Deng, Hao Su, Jonathan Krause, Sanjeev Satheesh, Sean Ma,
  Zhiheng Huang, Andrej Karpathy, Aditya Khosla, Michael Bernstein,
  Alexander~C. Berg, and Li~Fei-Fei.
\newblock {ImageNet Large Scale Visual Recognition Challenge}.
\newblock \emph{International Journal of Computer Vision (IJCV)}, 115\penalty0
  (3):\penalty0 211--252, 2015.
\newblock \doi{10.1007/s11263-015-0816-y}.

\bibitem[Sutskever et~al.(2014)Sutskever, Vinyals, and
  Le]{sutskever2014sequence}
Ilya Sutskever, Oriol Vinyals, and Quoc~V Le.
\newblock Sequence to sequence learning with neural networks.
\newblock In \emph{Advances in neural information processing systems}, pages
  3104--3112, 2014.

\bibitem[Teney et~al.(2018)Teney, Anderson, He, and van~den
  Hengel]{Teney_2018_CVPR}
Damien Teney, Peter Anderson, Xiaodong He, and Anton van~den Hengel.
\newblock Tips and tricks for visual question answering: Learnings from the
  2017 challenge.
\newblock In \emph{The IEEE Conference on Computer Vision and Pattern
  Recognition (CVPR)}, June 2018.

\bibitem[Vaswani et~al.(2017)Vaswani, Shazeer, Parmar, Uszkoreit, Jones, Gomez,
  Kaiser, and Polosukhin]{transformer}
Ashish Vaswani, Noam Shazeer, Niki Parmar, Jakob Uszkoreit, Llion Jones,
  Aidan~N. Gomez, Lukasz Kaiser, and Illia Polosukhin.
\newblock Attention is all you need.
\newblock \emph{CoRR}, abs/1706.03762, 2017.

\bibitem[Vedantam et~al.(2015)Vedantam, Lawrence~Zitnick, and
  Parikh]{vedantam2015cider}
Ramakrishna Vedantam, C~Lawrence~Zitnick, and Devi Parikh.
\newblock Cider: Consensus-based image description evaluation.
\newblock In \emph{Proceedings of the IEEE conference on computer vision and
  pattern recognition}, pages 4566--4575, 2015.

\bibitem[Vinyals et~al.(2015)Vinyals, Toshev, Bengio, and
  Erhan]{vinyals2015show}
Oriol Vinyals, Alexander Toshev, Samy Bengio, and Dumitru Erhan.
\newblock Show and tell: A neural image caption generator.
\newblock In \emph{Computer Vision and Pattern Recognition (CVPR), 2015 IEEE
  Conference on}, pages 3156--3164. IEEE, 2015.

\bibitem[Xu et~al.(2015)Xu, Ba, Kiros, Cho, Courville, Salakhudinov, Zemel, and
  Bengio]{xu2015show}
Kelvin Xu, Jimmy Ba, Ryan Kiros, Kyunghyun Cho, Aaron Courville, Ruslan
  Salakhudinov, Rich Zemel, and Yoshua Bengio.
\newblock Show, attend and tell: Neural image caption generation with visual
  attention.
\newblock In \emph{International Conference on Machine Learning}, pages
  2048--2057, 2015.

\bibitem[Yao et~al.(2017)Yao, Pan, Li, Qiu, and Mei]{yao2017boosting}
Ting Yao, Yingwei Pan, Yehao Li, Zhaofan Qiu, and Tao Mei.
\newblock Boosting image captioning with attributes.
\newblock In \emph{Proceedings of the IEEE Conference on Computer Vision and
  Pattern Recognition}, pages 4894--4902, 2017.

\bibitem[You et~al.(2016)You, Jin, Wang, Fang, and Luo]{you2016image}
Quanzeng You, Hailin Jin, Zhaowen Wang, Chen Fang, and Jiebo Luo.
\newblock Image captioning with semantic attention.
\newblock In \emph{Proceedings of the IEEE Conference on Computer Vision and
  Pattern Recognition}, pages 4651--4659, 2016.

\bibitem[Young et~al.(2014)Young, Lai, Hodosh, and Hockenmaier]{young2014image}
Peter Young, Alice Lai, Micah Hodosh, and Julia Hockenmaier.
\newblock From image descriptions to visual denotations: New similarity metrics
  for semantic inference over event descriptions.
\newblock \emph{Transactions of the Association for Computational Linguistics},
  2:\penalty0 67--78, 2014.

\end{thebibliography}

\end{document}